\documentclass[preprint,12pt,authoryear]{elsarticle}
\usepackage{amssymb}
\usepackage{cancel}
\journal{Expert Systems with Applications}

\usepackage{amsmath}
\usepackage{url}
\usepackage{xcolor}
\usepackage{apalike}
\usepackage{hyperref}
\usepackage{rotating}
\usepackage{afterpage}
\usepackage{float}

\biboptions{authoryear}

\begin{document}

\begin{frontmatter}



\title{Movie Trailer Genre Classification Using Multimodal Pretrained Features\fnref{ack2}}

\author[label1]{Serkan Sulun\corref{cor1}\fnref{ack1}}
\cortext[cor1]{Corresponding author.}
\fntext[ack1]{Serkan Sulun received the support of fellowship FCT - Fundação para a Ciência e a Tecnologia with the fellowship code 2022.09594.BD.}
\fntext[ack2]{This work has been funded by National Funds through the Portuguese funding agency, FCT - Fundação para a Ciência e a Tecnologia, within project LA/P/0063/2020.}
\ead{serkan.sulun@inesctec.pt}
\author[label1,label2]{Paula Viana}
\ead{pmv@isep.ipp.pt, paula.viana@inesctec.pt}
\author[label3]{Matthew ~E.~P.~Davies}
\ead{matthew.davies@siriusxm.com}

\affiliation[label1]{organization={Institute for Systems and Computer Engineering, Technology and Science (INESC TEC)},
            city={Porto},
            postcode={4200-465},
            country={Portugal}}

\affiliation[label2]{organization={ISEP,Polytechnic of Porto, School of Engineering},
            city={Porto},
            postcode={4200-072},
            country={Portugal}}

\affiliation[label3]{organization={SiriusXM},
            city={New York},
            postcode={10020}, 
            state={NY},
            country={USA}}

\begin{abstract}
We introduce a novel method for movie genre classification, capitalizing on a diverse set of readily accessible pretrained models. These models extract high-level features related to visual scenery, objects, characters, text, speech, music, and audio effects. To intelligently fuse these pretrained features, we train small classifier models with low time and memory requirements. Employing the transformer model, our approach utilizes all video and audio frames of movie trailers without performing any temporal pooling, efficiently exploiting the correspondence between all elements, as opposed to the fixed and low number of frames typically used by traditional methods. Our approach fuses features originating from different tasks and modalities, with different dimensionalities, different temporal lengths, and complex dependencies as opposed to current approaches. Our method outperforms state-of-the-art movie genre classification models in terms of precision, recall, and mean average precision (mAP). To foster future research, we make the pretrained features for the entire MovieNet dataset, along with our genre classification code and the trained models, publicly available.
\end{abstract}




\end{frontmatter}

\section{Introduction}
\label{sec:introduction}

The classification of movies into genres serves as a vital navigational tool in the immense catalog of cinematic content, allowing audiences to locate films aligned with their tastes and interests. It also provides a framework for scholars to explore thematic patterns, the evolution of narratives, and cultural shifts reflected in cinematic art. Moreover, for industry stakeholders, accurate genre classification guides key decisions around marketing and distribution strategies. The cinematic world contains an extraordinary blend of visual and auditory stimuli, each film constituting a unique assortment of aesthetics, themes, and narrative structures. However, this incredible diversity makes it challenging to accurately categorize films by genre.

Movie trailers serve as condensed representations of the full-length film, encompassing an array of vital elements within a brief span. Trailers provide a comprehensive overview of a film's narrative and genre by incorporating pivotal scenes, characters, objects, and vital auditory cues such as dialogue, music, and informative text elements. \textcolor{black}{The availability of movie trailers on online video platforms enables researchers to link movie metadata such as genre to the URL of the movie trailers, creating large datasets for movie trailer genre classification such as MovieNet \citep{movienet} and Moviescope \citep{moviescope}.}


\textcolor{black}{In trailer genre classification, as well as in almost all video classification tasks, deep neural networks (DNNs) constitute the state-of-the-art.} The majority of these models use raw data such as pixels and audio waveforms as their inputs. In contrast to older methods that use hand-crafted features, DNNs are expected to extract useful high-level features from raw data in an implicit way within the model layers. \textcolor{black}{These models often contain millions of parameters, requiring significant amounts of time, memory, and training data. Specifically in video classification, the uncompressed size of the raw video frame sequence increases the time and memory requirements even further. Previous works deal with this problem by processing only a fixed and small number of input frames or clips \citep{movienet, wehrmann}.} However, this approach inherently leads to information loss, potentially compromising classification accuracy. \textcolor{black}{Other works tackle this issue by using lower-dimensional, non-visual, and non-auditory inputs such as movie posters, plots, and metadata, for movie genre classification \citep{trailer_metadata_classification, trailer_poster_classification, trailer_poster_classification_2}. These approaches are dataset-specific and rely on auxiliary information, failing to classify solely based on video elements, namely on raw video pixels and audio waveforms.}

\textcolor{black}{The computational complexity of processing raw videos increases further with the adaptation of larger models---a trend that continues to expand over the years \citep{larger1,larger2}. When the training data is limited, using larger input dimensionality and larger models also increases the chance of overfitting \citep{overfitting}. The common and very similar ways to deal with these issues are transfer learning, fine-tuning, and using pretrained features \citep{transfer1,transfer2}. These terminologies can be distinguished as follows: Transfer learning involves applying a model trained on one task to a different task. Fine-tuning entails further training the transferred model for the new task. Using pretrained features involves performing inference on the data using a pretrained model, and then feeding the resulting outputs or activations (features) into a new model that is trained.} \textcolor{black}{These methods are frequently employed in video processing tasks such as human activity recognition, video summarization, and video recommendation \citep{activity_e, summarization_e, recommendation_e}.}

\textcolor{black}{Using pretrained features brings multiple advantages. Firstly, since the pretrained model is used in inference mode, its weights remain frozen, therefore reducing the time and memory complexity by requiring only a forward pass, without an additional backward pass. Secondly, it considerably reduces the size of the input fed to the subsequent model that is being trained. As an example, a very small video frame contains $224 \times 224 \times 3 = 163{,}968$ values, while the state-of-the-art image analysis model CLIP (Contrastive Language-Image Pre-Training) represents an image using only $512$ values \citep{clip}. This size reduction becomes more important as we use trailer videos that average around $2$ minutes in length, containing a long sequence of frames. Thirdly, assuming that the pretrained features of multimedia analysis models present valuable and relevant information for the task of video classification, their use removes the need to train new models to extract similar features implicitly from raw data. This reduces the size of the subsequent model that needs to be trained. Finally, since it reduces the input size and the number of weights in training, it also reduces the chance of overfitting \citep{overfitting}.}

\textcolor{black}{In this work, we focus on classifying movie trailers into genres using pretrained features that we extract entirely from video and audio. To obtain the relevant pretrained features, we exploit the availability of open-source pretrained multimedia analysis models.} Specifically, we employ image analysis, optical character recognition, automatic speech recognition, audio tagging, and music classification models to incorporate information about visual scenery, objects, text, speech, music, and audio effects. These pretrained models are used in inference mode, meaning that they are not fine-tuned, resulting in much lower time and memory requirements. \textcolor{black}{Another important strength of our methods is the use of the transformer model to fuse and process the pretrained features and predict the genre of the input video \citep{transformer}. The transformer model is designed to process long input sequences efficiently while handling long-term dependencies. In our case, we feed the pretrained features that stem from many video frames and audio chunks into the transformer. The small size of the pretrained features and the efficiency of the transformer model enables us to use entire videos while maximizing the available information, and not use a fixed and small number of input frames or audio chunks. Furthermore, since the pretrained models extract meaningful features from raw data, using a shallow transformer with one or two layers as the classifier becomes sufficient. We propose and compare multiple architectures to handle the non-trivial task of merging and processing multimodal pretrained feature sequences with complex correspondences and different temporal lengths and dimensionalities.} 

\textcolor{black}{We implement additional models to reflect some of the works in the literature and to compare against our methods. As done by \cite{movienet}, we use a temporal averaging module followed by a multi-layer perceptron (MLP) instead of the transformer, to average the pretrained features along the time dimension, and then classify them. Our empirical results show that the transformer outperforms this approach, especially when more input frames are used. We secondly implement a baseline model that works on raw video and audio, to mirror some of the recent models used in trailer genre classification \citep{trailer_metadata_classification}. While we empirically show that this approach performs worse than our proposed models, our preliminary experiments also show that it is prone to overfitting when trained end-to-end. Lastly, we compare our results with other models addressing trailer genre classification on the MovieNet dataset, demonstrating a notable performance improvement.} 


\textcolor{black}{We summarize the shortcomings of the previous works in trailer genre classification and how we address them as the following:
\begin{itemize}
    \item The complexity introduced by the large input space due to the use of raw video and audio: By using pretrained features instead of raw pixels and waveforms, we significantly reduce input complexity, dimensionality, and effectively, the chance of overfitting.
    \item The information loss caused by using a low and fixed number of input frames: Alongside using inputs with lower dimensionality, we utilize the transformer architecture that can process the sequences from the videos in their entirety while handling long-term correspondences. 
    \item Inability to classify solely using videos due to dependence on dataset-specific content such as poster, plot, or metadata: The pretrained features that we used are obtained entirely from video and audio. As a result, our models can theoretically be used for any video classification task.
\end{itemize}
}

In summary, our contributions are as follows:

\begin{itemize}
\item We propose new strategies for leveraging a variety of pretrained features as inputs and processing them using shallow neural networks for classification capable of handling different input sizes in both channel and temporal dimensions.

\item By utilizing the transformer model, we leverage all video keyframes and the entire audio, effectively handling videos of varying lengths without the need for \textcolor{black}{picking out a constant number of frames or average pooling}. We also quantitatively demonstrate that the classification performance improves steadily with the inclusion of more frames.

\item We improve the state-of-the-art for genre classification significantly, on the largest movie genre classification dataset, namely MovieNet.

\item We make available the pretrained features for the MovieNet dataset, as well as our code and trained models.\footnote{ \url{https://github.com/serkansulun/trailer-genre-classification} \\ \url{https://zenodo.org/records/13909366}}
\end{itemize}

\section{Related work}
\label{sec:related_work}

In this section, we first explain the cinematic datasets for movie genre classification and further motivate why we have chosen to work with the MovieNet dataset \citep{movienet}, and then we discuss the models specifically focused on movie genre classification.

\subsection{Cinematic datasets}

MovieLens is one of the earliest cinematic datasets, containing movie ratings and metadata like genre and title \citep{movielens}. While the latest version includes $86k$ movies, it does not include direct links to the movie trailers. The trailers can only be obtained through crawling the Internet Movie Database (IMDb) website\footnote{\url{https://www.imdb.com}}, which conflicts with their terms of use. \textcolor{black}{The MM-IMDb (Multimodal IMDb) dataset merges movie posters with the metadata obtained from MovieLens, resulting in $26k$ movies with posters, plots, genres, and other metadata, while lacking trailers or any types of videos \citep{mm-imdb}.} The LMTD (Large Movie Trailer Dataset) is a collection of features and metadata from $3500$ of movie trailers, although the project is currently discontinued and the data is unavailable \citep{lmtd}. MovieScope is a comprehensive dataset for multi-modal movie analysis, including data like movie trailers, posters, plot synopses, user reviews, and visual-auditory features, belonging to $5k$ distinct movies \citep{moviescope}. The MMTF-14K (Multimodal Movie Trailer Features dataset) provides multimodal features extracted from $14k$ movie trailers and their metadata such as user reviews and genre \citep{mmtf}. The Condensed Movies dataset includes full individual scenes, rather than trailers, alongside the plot and the characters, from $4k$ movies \citep{condensed}. 

MovieNet is a large-scale, holistic dataset providing movie, trailer, poster, subtitle, plot, tags, and metadata including genre \citep{movienet}. While it contains metadata belonging to $375k$ movies, trailers are available via YouTube links for $33k$ movies, making MovieNet the largest cinematic dataset in terms of both metadata and trailers. The work also includes the genre classification performance of several state-of-the-art video classification models, providing strong baselines to compare against our work. Overall, because MovieNet is the largest movie trailer dataset, has direct links to trailers, and has quantitative genre classification results to serve as a baseline, we trained and evaluated our models on this particular dataset.

\subsection{Movie genre classification}

One of the earliest works on movie genre classification obtains keyframes using scene detectors, and extracts hand-crafted visual features such as roughness, ruggedness, and openness, on a privately collected dataset \citep{zhou}. Genre classification is achieved by comparing the distance between the feature vectors from the training and testing sets. 

One of the first works that use neural networks for movie genre classification also utilizes visual pretrained features \citep{wehrmann}, using the LMTD dataset \citep{lmtd}. They use pretrained features of classification models that were trained on ImageNet \citep{imagenet} and Places365 \citep{places} datasets, and audio spectrograms. The features from individual frames are fused using a convolution-through-time module, which can be thought of as a standard convolutional neural network (CNN) where the kernel length is equal to the input feature length of each keyframe and the kernel traverses the features belonging to subsequent frames, along the temporal dimension \citep{wehrmann}. The kernel length along the temporal dimension is $3$, meaning that the model can exploit the correspondence between only $3$ frames. Moreover, since the input and output of the CNN have varying numbers of frames, they further take the maximum along the temporal dimension to have a fixed-size vector to feed into the final classification layer. This inevitably leads to a loss of information. Another work exploits the correspondence between facial emotions and cinematic genre, by first extracting human faces from the trailer videos, then classifying their emotions, and finally mapping the emotions to the cinematic genres \citep{yadav}. 

The work presenting the MovieNet dataset also introduces a model for movie genre classification, alongside the results from other state-of-the-art video classification models \citep{movienet}. While the model they introduce does surpass the performance of the state-of-the-art models, it only uses $8$ clips, each with $3$ frames from the entire trailer. During the inference phase, predictions are made for each individual clip, and these are subsequently averaged to generate the final prediction. This approach, however, fails to account for the long-term correspondence that exists within the trailers. Finally, the MovieCLIP model first trains a scene classification model and then uses its final activations to feed into an additional genre classification model, specifically working with the Moviescope dataset \citep{movieclip}.

\textcolor{black}{Some recent works have used transformers for movie genre classification. \citet{trailer_metadata_classification} used transformers to individually process raw video frames and raw audio, and later fused the resulting activations with metadata information and poster. \citet{trailer_poster_classification_2} and \cite{trailer_poster_classification} used pretrained transformers to process movie posters and plots, excluding any use of videos, for genre classification on the MM-IMDb dataset \citep{mm-imdb}. Our work differs from these recent approaches by using pretrained features instead of raw inputs; and by working solely on video and audio, namely, only using raw video pixels and audio waveforms. We also implement a baseline model that is similar to \cite{trailer_metadata_classification} which uses raw inputs and we empirically show that it performs worse than our proposed models.}

\section{Methodology}
\label{sec:methodology}

Our overall task is to classify each cinematic trailer into its corresponding genres. While the methods in the literature use raw pixels and audio as the input to the classifier model, we exploit learned features created by existing pretrained deep neural networks (DNNs). Since these features are already created by deep and powerful models, we can train shallow models to process these features and predict the final output, enabling us to cut down on the training time and resources, while still making use of semantically relevant features.

\subsection{Creating training data}

We used the YouTube trailers in the MovieNet dataset \citep{movienet}, choosing the videos with the lowest resolution where the height is at least $300$ pixels. We re-encoded the videos using the x265 encoder with a constant rate factor of $23$. Using FFmpeg \citep{ffmpeg}, we detected the scene boundaries and extracted the frames that are exactly in the middle of two scene boundaries. We also re-encoded the audio using the Opus codec, with a bitrate of $48k$ in mono. For each video, we extract various features using pretrained models and store them so that we can simply load them while training the classification model. We use the genre(s) of each movie from the MovieNet dataset as the ground-truth. Importantly, this dataset is multi-label, indicating that an individual video can be associated with multiple genres.

\subsubsection{Feature extraction}

\begin{figure}[t]
\centering
\includegraphics[width=0.5\columnwidth]{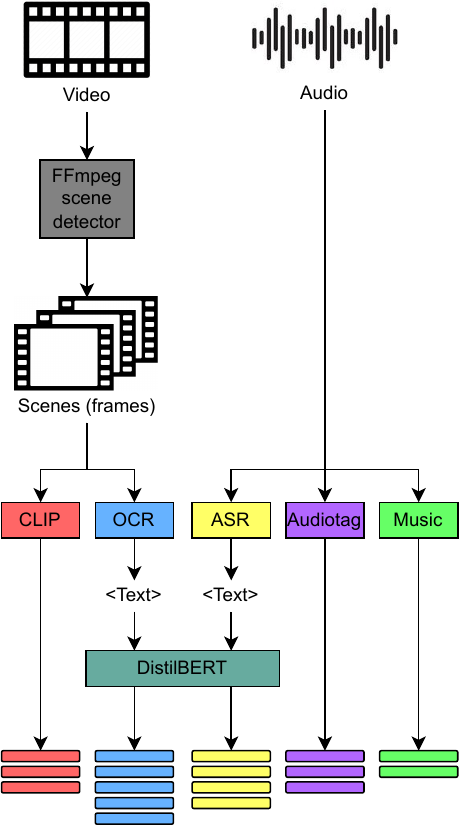} %
\caption{Feature extraction pipeline.}
\label{fig:features}
\end{figure}

The overall feature extraction pipeline is shown in Figure \ref{fig:features}, and each feature is explained below.

\paragraph{CLIP}
Contrastive Language-Image Pretraining (CLIP), is the state-of-the-art model for image understanding, that was pretrained using contrastive learning, using a large number of images with captions available on the internet \citep{clip}. The pretrained model first resizes the image where the longer side has $224$ pixels and further takes a $224\times224$ center crop. Afterward, the pretrained model encodes each frame to a vector with a length of $512$.

\paragraph{Audiotag}
To extract audio features, we employed a model that was pretrained on the task of audio tagging \citep{audiotag}. This model operates on chunks of $3$ seconds and outputs a vector of probabilities for $527$ different labels. Since we are interested in encoding the audio into feature vectors, we don't use the final classification layer and extract the activations prior to that, ending up with vectors with a length of $128$, for each audio chunk.

\paragraph{Musicnet}
Since we are dealing with cinematic trailers, the soundtrack is an important element that further needs investigation. Hence we extract another audio-related feature, namely the musical feature, using a model that was pretrained on the task of music genre classification \citep{musicnet}. This model operates on chunks of $22$ seconds and outputs a vector of probabilities for $50$ different labels. Similarly, we extract the activations prior to the final layer, ending up with vectors with a length of $64$, for each audio chunk.

\paragraph{OCR}
We also performed optical character recognition (OCR) on each frame using the PaddleOCR model \citep{paddle}. We additionally used a pretrained spell correction model on the produced output \citep{spell_correction}. The overall output of this stage is in text format.

\paragraph{ASR}
We performed automatic speech recognition (ASR) on the audio using OpenAI's Whisper model \citep{whisper}. Since this is a learned model that was trained on natural text, there is no need for post-spell correction. However, on the output of Whisper, we used a pretrained language detection model to filter out non-English text \citep{language_detection}. When the output of ASR was in a non-English language, we simply replaced the entire text to denote the language, e.g. changing it to ``German language". Similarly, the output of this model is in text format.

\paragraph{DistilBERT}
We encode the output text of the OCR and ASR models using a pretrained language model, specifically DistilBERT \citep{distilbert}, which is a condensed and compressed variant of the BERT (Bidirectional Encoder Representations from Transformers) model \citep{bert}, achieved through knowledge distillation \citep{distillation_2006,distillation_2015}. DistilBERT utilizes fewer layers than BERT and learns from BERT's outputs to mimic its behavior. This model converts the input text into tokens of words and sub-words, and encodes each one of them as a vector with a length of $768$.

\subsection{Classification}

We built multiple models to take the previously extracted features of a given video as input and predict the genre of the video. Combining different features is challenging, especially since the encoded vectors have different lengths in both channel and temporal dimensions. For example, considering the audio event and music features, the lengths of each vector are $128$ and $64$, respectively. Furthermore, since the audio event and music networks operate on chunks of audio with lengths of $3$ and $22$ seconds respectively, for the same video, there are more embedded vectors for the former. And finally, the number of vectors for the same feature differs between different videos. We addressed these issues using different solutions and different classification models.

We built and trained three distinct models: \textit{a multi-layer perceptron (MLP)}; the \textit{single-transformer} model that integrates features across all modalities; and the \textit{multi-transformer} model, where individual transformers handle features from specific modalities.

The final layer of all our models is a fully-connected (FC) layer with a size of $21$, outputting probabilities belonging to $21$ different genres. This layer is followed by a sigmoid layer to make sure each output is a probability between $0$ and $1$. We note that predicted probabilities do not add up to $1$ due to the multi-label setting. 

\subsubsection{MLP}

\begin{figure}[t]
\centering
\includegraphics[width=0.75\columnwidth]{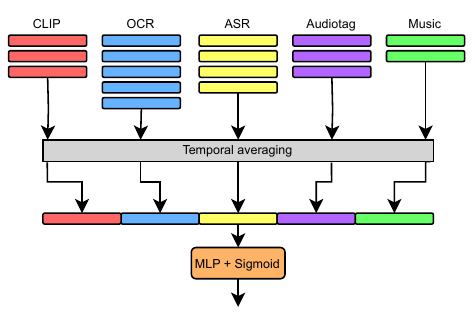} %
\caption{Multilayer perception (MLP). The last layer of the MLP block is the classification layer.}
\label{fig:mlp}
\end{figure}

We first implemented a simple MLP classifier \textcolor{black}{to reflect the state-of-the-art method used by \cite{movienet}. We note that \cite{movienet} used raw video and audio from a few small segments of video, while we use pretrained features stemming from the entire video. Our model}  can be seen in Figure \ref{fig:mlp}. Since MLPs require a fixed-length input, we averaged the feature vectors from each modality along the temporal dimension \textcolor{black}{similar to \cite{movienet}}. We then concatenated the averaged vectors from different modalities and fed them into the MLP.

\begin{figure}[th]
\centering
`\includegraphics[width=0.75\columnwidth]{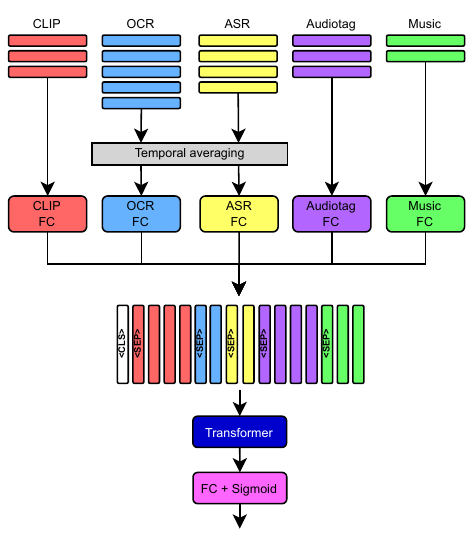} %
\caption{Single-transformer. Regarding the transformer's output, only the output vector that corresponds to the $<$CLS$>$ vector is passed to the final classification layer. The $<$SEP$>$ vectors are used to explicitly separate the elements of each modality.}
\label{fig:single}
\end{figure}

\begin{figure}[th]
\centering
\includegraphics[width=0.75\columnwidth]{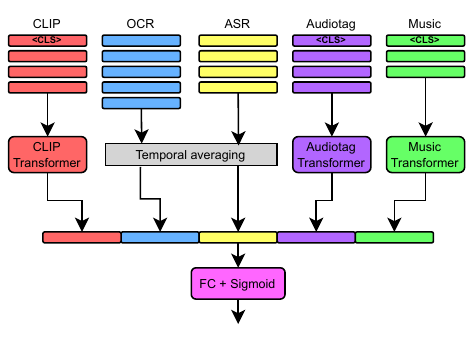} %
\caption{Multi-transformer. Regarding the output of each transformer, only the output vectors that correspond to the $<$CLS$>$ vectors are used. For inputs that are averaged along the temporal dimensions, no $<$CLS$>$ vectors are used since temporal averaging already yields a single vector.}
\label{fig:multi}
\end{figure}

\subsubsection{Single-transformer}

To address the significant information loss caused by averaging features along the temporal dimension, we utilized a transformer model to leverage short- and long-term correspondences within the video sequence. The transformer is a state-of-the-art sequence processing model, capable of efficiently handling long sequences through its built-in multi-headed attention mechanism \citep{transformer}. We employed the transformer as a sequence classifier by prepending the sequences with a special learnable vector, called the $<$CLS$>$ vector. While the transformer generates output vectors for each element of the input sequence, in classification tasks, only the output vector corresponding to the $<$CLS$>$ vector is forwarded to the next layer, while the others are discarded \citep{bert}.

While the transformer can handle sequences with varying lengths, the vectors in the sequence still need to have equal sizes along the channel dimension. To ensure that, we used fully-connected (FC) layers to transform feature vectors from different modalities into the same size. We note that this layer is applied to each vector in the sequence individually, hence not changing the number of vectors in its input sequence. Next, we concatenated all the vectors along the time dimension before feeding them into a single transformer. \textcolor{black}{This approach aims at exploiting the correspondences between different features at the temporal level by processing all of them as a single sequence. However, a trade-off is the complexity of handling different modalities with a single transformer block using a single set of weights. The overall model is shown in Figure \ref{fig:single}}.

To make sure the transformer can distinguish vectors from different modalities, we used separately learned positional embeddings for sequences from each modality \citep{learned_position}. We additionally made use of separator ($<$SEP$>$) vectors, namely, an encoded version of the $<$SEP$>$ token \citep{bert}. These are learned vectors that are specific for each modality, prepended to each sequence. Finally, as in almost all classifier transformers, we prepended the input sequence to the transformer with a learnable classifying ($<$CLS$>$) vector, namely, an encoded version of the $<$CLS$>$ token \citep{bert}. The output vector corresponding to this classifying vector, i.e. the first vector, is considered as the prediction of the transformer and fed into the final fully-connected layer, while the remaining vectors of the transformer's output sequence are discarded.

\subsubsection{Multi-transformer}

With our hypothesis that using a single transformer to process features from multiple modalities can be inefficient \textcolor{black}{due to the increased complexity of the input}, we devised our final model that incorporates different transformer models to process features from different modalities, as shown in Figure \ref{fig:multi}. The inputs of all transformers are prepended with $<$CLS$>$ token vectors, and the corresponding output vectors are concatenated in channel dimension to be fed into the fully-connected layer to obtain the final probabilities. \textcolor{black}{This approach utilizes the potential correspondences between different features at the global level, though not at the temporal level. }

\subsubsection{Optional temporal averaging}

We optionally averaged the text-related features such as OCR and ASR, along the temporal dimension. Non-textual features, namely CLIP, Audiotag, and Musicnet are obtained from the activations \textit{before} the final prediction layer. Whereas the OCR and ASR features stem from running the DistilBERT model on the predicted text. Here, any error in the predicted text is propagated into the DistilBERT model, corrupting the output features. In order to reduce the resulting noise, we experimented with averaging the textual features, namely OCR and ASR, along the temporal dimension.

\subsubsection{Vision transformer baseline model}

We also implemented a strong baseline model employing state-of-the-art vision transformers (ViT) that work on sequences of 2-dimensional raw visual frames and audio spectrograms, as opposed to the previously introduced models that work with sequences of 1-dimensional pretrained features \citep{vit}. ViT segments the image or audio spectrogram into patches, encodes them, and then processes the sequence of encoded vectors using the standard transformer model. While it is purely attention-based, without any convolutional layers, it outperforms CNNs, achieving state-of-the-art results in image \citep{vit}, video \citep{vivit}, and audio \citep{ast} classification. 

While our primary objective is video classification, it is important to note that directly applying the video vision transformer (ViViT) \citep{vivit} to our task is not a suitable approach. There are two main reasons for this. First, ViViT doesn't incorporate audio, which is a crucial aspect of our task. Second, our dataset contains discontinuous frames that depict scenes with no visual continuity between them. As a result, it is more appropriate to handle these frames individually rather than concatenating patches from different frames, as ViViT does.

\begin{figure}[t]
\centering
\includegraphics[width=0.55\columnwidth]{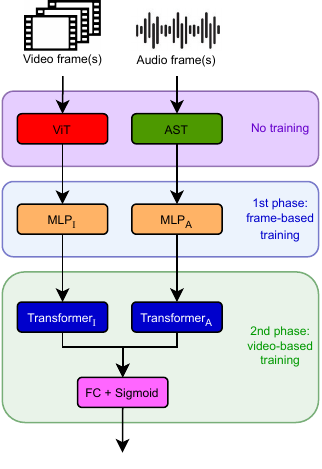} %
\caption{Baseline model that works with 2-dimensional raw video frames and audio (spectrogram) frames.} The subscripts ``I" and ``A" refer to image and audio, respectively.
\label{fig:baseline}
\end{figure}

We utilized a standard pretrained vision transformer (ViT) to process images and a modified version of ViT, known as an audio spectrogram transformer (AST), designed and pretrained specifically for handling audio spectrograms \citep{ast}, to process audio data. Our preliminary experiments showed that training the full model led to overfitting issues that couldn't be mitigated through standard techniques like dropout \citep{dropout}. To avoid this, we kept the parameters of ViT and AST frozen and replaced their output layers with trainable MLPs, naming them image-MLP and audio-MLP.

The pretrained AST is designed to work with audio spectrograms organized into 10-second segments, which we refer to as ``spectrogram frames". To distinguish them, we use the term ``image frames" to describe the visual scenes in our context. Since the trailers in our dataset are longer than $10$ seconds, we split the full audio spectrogram into $10$-second chunks, with a $50\%$ overlap. Similarly, the ViT works with individual image frames, while our trailers consist of multiple frames. This leads to an output sequence of vectors, where each vector corresponds to an individual image or audio frame. To process these sequences and obtain the final predictions, we developed a fusion module. To this end, we used standard transformer classifiers for both image and audio sequences, yielding a single vector for each modality. Finally, we concatenated these two vectors and processed them using a linear layer and a sigmoid layer to obtain the final label predictions.

Even when the parameters of pretrained ViT and AST were kept frozen, training the full model in an end-to-end fashion led to overfitting. This is because our training dataset is relatively small with only $33k$ samples, meanwhile, the original authors of ViT and AST trained their models with extensive datasets such as Imagenet with $21M$ samples \citep{imagenet} and AudioSet with $2M$ samples \citep{audioset}, respectively. To address this overfitting issue while reducing the training dataset size to 1\% and 0.1\% of the original ones, we employed a two-stage training approach. Initially, we trained the image- and audio-MLPs separately on individual frames. During this phase, the target was the label of the video to which the frames belonged. Following this, we further froze the parameters of the MLPs and trained the fusion module using complete videos. The baseline model and its training scheme can be seen in Figure \ref{fig:baseline}.

\subsubsection{Training details and hyperparameters}

We applied a filtering process to exclude extremely long and short videos from our dataset. Specifically, we computed the quartiles of video durations and excluded samples that were outside the inner fence, i.e., durations shorter than $Q1 - 1.5 \times IQR$ (corresponding to $19.6$ seconds) and longer than $Q3 + 1.5 \times IQR$ (corresponding to $214.4$ seconds). This filtering yielded $26412$ videos from the original set of $32647$. Moreover, following the approach in the MovieNet paper \citep{movienet}, we limited the labels to the $21$ most frequent genres. These labels are shown in Table \ref{table:genres}.

To ensure an unbiased split of the dataset, we arranged the videos alphabetically by their YouTube IDs---given that these IDs are generated randomly. Following the methodology in the original MovieNet paper, we divided the dataset into training, validation, and testing splits with ratios of $0.7$, $0.1$, and $0.2$, corresponding to $18488$, $2641$, and $5283$ samples respectively \citep{movienet}. Hyperparameters were determined using a grid search, optimizing for the highest mean average precision on the validation split. The test split was utilized solely for reporting the final results.

While a straightforward way to balance precision and recall is to change the decision threshold, this method does not allow for comparison against the works in the literature where most report their results while setting a fixed decision threshold at $0.5$ \citep{movienet}. Alternatively, during training, we can balance precision and recall by applying a constant weight to the loss associated with positive labels. For a single sample, the weighted binary cross-entropy loss becomes:

\begin{align*}
\textit{Loss} = -\frac{1}{C} \sum_{c=1}^{C} w \cdot y_c \cdot \log(p_c) + (1 - y_c) \cdot \log(1 - p_c)
\end{align*}

Here, $w$ represents the weight for positive labels and $C$ is the total number of classes. To align precision and recall values with those observed in MovieNet, we applied a weight of $0.25$.

Both the transformer architecture and the averaging module of the MLP architecture support sequences with varying lengths. However, during training, we employed fixed-length inputs to facilitate minibatch processing. Longer sequences were truncated to the desired length, while shorter ones were padded with zero-vectors. The length of feature vectors for each pretrained modality was individually determined through exploratory data analysis, specifically using box plots to analyze sequence lengths across all samples. The maximum sequence length was set to the upper adjacent value (upper whisker of the box plot), which is $1.5$ times the interquartile range above the third quartile. Consequently, the sequence lengths were defined as $216$ for CLIP, $64$ for OCR, $86$ for ASR, $140$ for Audiotag, and $18$ for Musicnet. When using a single-transformer model, the concatenated sequence length totaled $524$. It's important to note that during inference on single samples, our models can handle inputs with varying lengths without the need for padding or truncating.

For the MLP model, the number of layers is $1$, the model dimension is $256$, and the total number of parameters is $57k$. The single-transformer model has $2$ layers, $8$ attention heads, and a model dimension of $256$, totaling $8.56M$ parameters. The multi-transformer model has individual transformers each with $1$ layer, $8$ attention heads, and a model dimension of $128$, totaling $6.98M$ parameters.  
Considering the baseline model, both image-MLP and audio-MLP have $2$ layers and a dimensionality of $768$. The fusing transformers both have $1$ layer, $4$ attention heads and a dimensionality of $768$.
All transformers are trained with gradient clipping at a norm of $1$. 
All models are trained with Adam optimizer \citep{adam}, a learning rate of $1e-5$, a dropout rate of $0.5$, and a batch size of $32$. We implemented our models using the Pytorch library \citep{pytorch} and trained them on a single NVIDIA Quadro RTX 6000 GPU with $24$ GB memory.

\section{Experiments and results}

We first compare the performance of our models against the baseline model and the models presented in MovieNet paper \citep{movienet}. We performed inference on the test set one sample (video) at a time, using the full duration of each video without padding or truncating the sequences. Following the literature, we report macro-averaged precision, recall, and mean average precision (mAP) values, averaged over individual labels (genres).

Adjusting the precision-recall trade-off can be achieved by altering the decision threshold in inference or by modifying the weight assigned to positive labels during training. The MovieNet paper specifies a decision threshold of $0.5$ and does not make any reference to label weighting. We kept the decision threshold at $0.5$ and aimed for performance similar to the MovieNet paper by assigning a weight of $0.25$ to positive (existing) labels. 

Table \ref{table:overall} displays the overall performance of our models against the models in the literature. Our baseline model outperforms the models in the literature in terms of recall and mean average precision. Our models using pretrained features outperform all other models across all metrics. Our best-performing model, the multi-transformer improves the state of the art by a large margin. It outperforms all other models in all metrics except precision, where it performs marginally worse than our MLP model.

\begin{table}[t]
\centering
\caption{State-of-the-art models vs. ours. P, R, and mAP denote precision, recall, and mean average precision respectively. For precision and recall, $0.5$ is used as the decision threshold. Baseline results are obtained from the paper that introduces the MovieNet dataset, where the authors trained different state-of-the-art models, in addition to their own model, on their own dataset \citep{movienet}. The full forms of the abbreviated model names are as follows. TSN: Temporal Segment Network \citep{tsn}, I3D: Two-Stream Inflated 3D ConvNets \citep{i3d}, TRN: Temporal Relation Network \citep{trn}.}

\begin{tabular}{lccc}
\multicolumn{1}{c}{Models} & P@0.5          & R@0.5          & mAP            \\ \hline \hline
    TSN   & 78.31 & 17.95 & 43.70 \\
    I3D   & 69.58 & 16.54 & 35.79 \\
    TRN   & 77.63 & 21.74 & 45.23 \\
    MovieNet & 79.74 & 24.97 & 46.88 \\ \hline
    Baseline & 73.78 & 32.17 & 59.73 \\ \hline
    MLP   & \textbf{82.05} & 33.51 & 63.16 \\
    Single-transformer & 81.00 & 37.11 & 65.09 \\
    Multi-transformer & 82.00 & \textbf{38.33} & \textbf{66.02} 
\end{tabular}
\label{table:overall}
\end{table}

\begin{sidewaystable}[p]
\centering
\caption{Results per genre and the macro averages. MovieNet (MN) \citep{movienet} vs baseline model vs our best model multi-transformer. P, R, and mAP denote precision, recall, and mean average precision respectively. For precision and recall, $0.5$ is used as the decision threshold. }

\begin{tabular}{l|ccc|ccc|ccc}
             & \multicolumn{3}{c|}{P@0.5}               & \multicolumn{3}{c|}{R@0.5}                     & \multicolumn{3}{c}{mAP}     \\ \hline
            & MN            & Baseline & Ours          & MN             & Baseline      & Ours          & MN            & Baseline & Ours          \\ \hline
    Action & 73.96 & 86.59 & \textbf{87.37} & 22.21 & 40.48 & \textbf{48.12} & 54.60 & 75.64 & \textbf{79.20} \\
    Adventure & 75.24 & 74.71 & \textbf{77.51} & 24.72 & 22.48 & \textbf{28.67} & 53.06 & 56.76 & \textbf{60.93} \\
    Animation & 93.16 & \textbf{96.38} & 96.19 & 74.09 & 86.59 & \textbf{92.28} & 86.45 & 94.65 & \textbf{96.18} \\
    Biography & \textbf{100.00} & 53.85 & 70.00 & 0.04  & 2.86  & \textbf{5.71} & 9.13  & 24.29 & \textbf{36.68} \\
    Comedy & 68.61 & 88.55 & \textbf{90.39} & 48.65 & \textbf{57.94} & 57.05 & 68.81 & 85.47 & \textbf{87.84} \\
    Crime & 74.12 & 72.00 & \textbf{80.22} & \textbf{39.30} & 8.33  & 22.53 & 49.25 & 50.81 & \textbf{60.25} \\
    Documentary & 85.49 & 91.30 & \textbf{92.37} & 4.79  & 70.87 & \textbf{81.71} & 21.03 & 90.29 & \textbf{94.64} \\
    Drama & 71.16 & 85.70 & \textbf{89.40} & \textbf{79.42} & 49.16 & 55.06 & 79.95 & 83.08 & \textbf{86.77} \\
    Family & 82.55 & 83.08 & \textbf{86.58} & 27.11 & 40.14 & \textbf{48.08} & 52.19 & 67.09 & \textbf{73.60} \\
    Fantasy & 69.83 & 79.37 & \textbf{87.88} & 13.51 & 12.02 & \textbf{20.91} & 39.12 & 48.78 & \textbf{59.59} \\
    History & 82.90 & 63.16 & \textbf{90.00} & \textbf{12.52} & 6.49  & 9.73  & 34.41 & 33.63 & \textbf{41.05} \\
    Horror & 70.03 & 86.49 & \textbf{88.38} & 8.76  & 51.04 & \textbf{55.82} & 35.51 & 79.48 & \textbf{84.24} \\
    Music & \textbf{89.04} & 66.67 & 68.06 & 27.24 & 20.48 & \textbf{29.52} & 47.13 & 42.17 & \textbf{48.36} \\
    Musical & 73.58 & 70.59 & \textbf{75.00} & 4.45  & 11.21 & \textbf{14.02} & 22.88 & 32.86 & \textbf{41.27} \\
    Mystery & \textbf{76.42} & 0.00  & 57.14 & \textbf{7.76} & 0.00  & 1.20  & \textbf{39.70} & 25.28 & 30.68 \\
    Romance & 71.93 & 76.47 & \textbf{82.32} & 14.02 & 13.65 & \textbf{15.75} & 49.27 & 51.16 & \textbf{58.59} \\
    Sci-Fi & 81.35 & 72.09 & \textbf{83.60} & 14.51 & 29.67 & \textbf{37.80} & 44.14 & 54.11 & \textbf{68.00} \\
    Sport & \textbf{94.97} & 69.51 & 75.86 & 21.99 & 42.86 & \textbf{49.62} & 39.59 & 60.44 & \textbf{66.80} \\
    Thriller & 64.98 & \textbf{79.35} & 78.17 & 14.50 & 28.79 & \textbf{37.82} & 49.80 & 69.54 & \textbf{71.76} \\
    War   & \textbf{86.27} & 64.71 & 75.86 & 12.80 & 20.89 & \textbf{27.85} & 34.41 & 47.66 & \textbf{56.40} \\
    Western & 88.89 & 88.89 & \textbf{89.80} & 51.93 & 59.70 & \textbf{65.67} & 73.99 & 81.14 & \textbf{83.53} \\ \hline
    AVERAGE & 79.74 & 73.78 & \textbf{82.00} & 24.97 & 32.17 & \textbf{38.33} & 46.88 & 59.73 & \textbf{66.02} \\ 

\end{tabular}

\label{table:genres}
\end{sidewaystable}

In Table \ref{table:genres} we compare the performance of our best model multi-transformer, our baseline and the MovieNet model \citep{movienet}, across all genres. Out of all $21$ genres, our model outperforms the other models in $14$ of them in terms of precision, $16$ of them in terms of recall, and $20$ of them in terms of mean average precision.
Outlier values such as very low recall for genres such as Biography, History, and Mystery can be attributed to the inherent imbalance within the MovieNet dataset \citep{movienet}. We deliberately avoided balancing the data before training to maintain a fair comparison with the classification model presented in the MovieNet paper. We also believe that this dataset closely reflects the real-world distribution of cinematic genres, accurately portraying the relative rarity of genres such as Biography, History, and Mystery.

In Table \ref{table:features} we present an ablation study using our best-performing model multi-transformer, demonstrating the gain in terms of mean average precision along with the increase in runtime, due to the addition of each feature while comparing against our baseline model. Runtime is defined as the average duration in seconds, to process a single video, both extracting its pretrained features and classifying it, during inference. Although the inclusion of textual features such as OCR and ASR initially seems to hurt the performance, possibly due to the text prediction errors, incorporating their averaged version along the temporal dimension reduces the noise and does improve mean average precision. Incorporating textual features also significantly increases runtime. This is primarily because models such as CLIP, Musicnet, and Audiotag are designed to predict a single embedding vector or label, while OCR and ASR models predict text sequences, which can be viewed as predicting many labels corresponding to each word and subword, in an autoregressive way. But most importantly, the settings in which the textual features are excluded yield a better classification performance and faster runtime compared to the baseline. Furthermore, our classification models can seamlessly incorporate features that result from newer and potentially more efficient ASR and OCR models that can be developed in the future, further reducing the overall runtime.

\begin{table}[t]
\centering
\caption{The effect of inclusion of pretrained features on mean average precision and runtime. The runtime is defined as the average time, measured in seconds, to classify a single video in inference mode, including the extraction of the pretrained features. The first row shows the performance of the baseline model. Asterix ($\ast$) indicates averaging the features over the sequence (temporal) dimension.}

\begin{tabular}{|c|c|c|c|c||c|c|c|}
\hline
CLIP & Musicnet & Audiotag & OCR & ASR & mAP & Runtime (s.)          \\ \hline \hline
              &               &               &               &               & 59.73 & 7.08 \\ \hline \hline
$\checkmark$  &               &               &               &               & 64.73 & 5.32     \\ \hline
$\checkmark$  & $\checkmark$  &               &               &               & 65.17 & 5.76      \\ \hline
$\checkmark$  & $\checkmark$  & $\checkmark$  &               &               & 65.31 & 5.95      \\ \hline
$\checkmark$  & $\checkmark$  & $\checkmark$  & $\checkmark$  &               & 63.33 & 25.85      \\ \hline
$\checkmark$  & $\checkmark$  & $\checkmark$  & $\ast$        &               & 65.46 & 31.52      \\ \hline
$\checkmark$  & $\checkmark$  & $\checkmark$  & $\checkmark$  & $\checkmark$  & 64.66 & 31.44      \\ \hline
$\checkmark$  & $\checkmark$  & $\checkmark$  & $\ast$        & $\ast$        & \textbf{66.02} & 33.59  \\ \hline
\end{tabular}

\label{table:features}
\end{table}

\begin{figure}[H]
\centering
\includegraphics{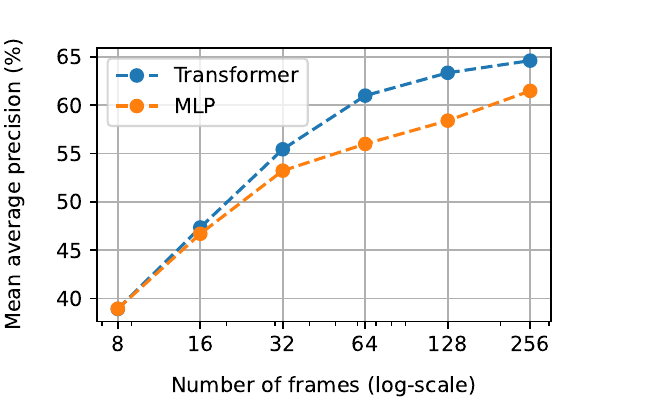}
\caption{Number of frames vs. mean average precision, only using CLIP features.}
\label{fig:frames}
\end{figure}

Finally in Figure \ref{fig:frames} we attempt to further explain the reasons behind the success of our models. The models in the literature extract a small and fixed number of frames at random locations from each video for classification. The MovieNet baseline \citep{movienet} only uses $8$ scenes. They furthermore use MLPs, which cannot process sequences, hence the features from these small number of frames need further averaging along the temporal dimension, causing additional loss of information. Using a simplified experiment, we show the importance of the number of frames fed into the model. Using only the CLIP features, we employ $8$, $16$, $32$, $64$, $128$ and $256$ frames obtained from random locations in each video. We only use the CLIP features since the number of feature vectors differs amongst the pretrained features, and finding an exact number of features that would correspond to the specific number of CLIP features is challenging. We process the features using an MLP and a transformer. Regarding the MLP, the features from different frames are averaged along the temporal dimension as in \citep{movienet}. On the contrary, the transformer model is capable of processing sequences hence we feed the features from each frame \textit{as is}.

In Figure \ref{fig:frames}, we can see that using a higher number of frames improves the performance significantly. Furthermore, as the number of frames increases, the superiority of the transformer over the MLP becomes apparent, due to its ability to process long sequences seamlessly.

In summary, the success of our models is due to using various pretrained features, using \textit{all} scenes as frames and the entire audio, and finally avoiding averaging features along the temporal dimension and feeding them all to the transformer that is the state-of-the-art model for processing long sequences.

\section{Conclusion}
\label{sec:conclusion}

In conclusion, our research presents a pioneering approach to movie genre classification, significantly enhancing the performance beyond current methodologies. Our strategy leverages a range of pretrained models to extract and intelligently fuse high-level features associated with visual scenery, characters, text, speech, music, and audio effects. A fundamental element of our approach is the employment of the transformer model, which facilitates the efficient handling of sequences of any length. Importantly, our method capitalizes on the entirety of the information present in all frames of movie trailers, contrasting with traditional models that are restricted to a fixed and low number of frames. \textcolor{black}{Finally, our method purely operates on video and audio, without requiring any dataset-specific auxiliary data, making it potentially applicable to any video classification task.}

Upon the acceptance of our paper, we will make the pretrained features for the entire MovieNet dataset publicly available, along with our genre classification code and trained models, to contribute to ongoing advancements in this field. \textcolor{black}{Even though our models seem complex, involving multiple pretrained models, we made sure all the components and their pretrained weights are packed into a single codebase in a ``plug-and-play" manner. We also include a script only for inference on single videos, helping non-academic users to analyze individual videos. These resources are expected to advance movie genre classification and multimedia analysis in general while benefiting both researchers and the broader public.}

\bibliography{references.bib}

\begin{thebibliography}{48}
\expandafter\ifx\csname natexlab\endcsname\relax\def\natexlab#1{#1}\fi
\providecommand{\url}[1]{\texttt{#1}}
\providecommand{\href}[2]{#2}
\providecommand{\path}[1]{#1}
\providecommand{\DOIprefix}{doi:}
\providecommand{\ArXivprefix}{arXiv:}
\providecommand{\URLprefix}{URL: }
\providecommand{\Pubmedprefix}{pmid:}
\providecommand{\doi}[1]{\href{http://dx.doi.org/#1}{\path{#1}}}
\providecommand{\Pubmed}[1]{\href{pmid:#1}{\path{#1}}}
\providecommand{\bibinfo}[2]{#2}
\ifx\xfnm\relax \def\xfnm[#1]{\unskip,\space#1}\fi
\bibitem[{Ak et~al.(2023)Ak, Lee, Xu \& Shen}]{trailer_poster_classification}
\bibinfo{author}{Ak, K.~E.}, \bibinfo{author}{Lee, G.-G.}, \bibinfo{author}{Xu,
  Y.}, \& \bibinfo{author}{Shen, M.} (\bibinfo{year}{2023}).
\newblock \bibinfo{title}{Leveraging efficient training and feature fusion in
  transformers for multimodal classification}.
\newblock In {\it \bibinfo{booktitle}{IEEE International Conference on Image
  Processing, ICIP 2023, Kuala Lumpur, Malaysia, October 8-11, 2023}\/} (pp.
  \bibinfo{pages}{1420--1424}).
\newblock \bibinfo{publisher}{IEEE}.
\newblock \DOIprefix\doi{10.1109/ICIP49359.2023.10223098}.
\bibitem[{Almeida et~al.(2022)Almeida, {de Villiers}, De~Freitas \&
  Velayudan}]{recommendation_e}
\bibinfo{author}{Almeida, A.}, \bibinfo{author}{{de Villiers}, J.~P.},
  \bibinfo{author}{De~Freitas, A.}, \& \bibinfo{author}{Velayudan, M.}
  (\bibinfo{year}{2022}).
\newblock \bibinfo{title}{The complementarity of a diverse range of deep
  learning features extracted from video content for video recommendation}.
\newblock {\it \bibinfo{journal}{Expert Systems with Applications}\/},  {\it
  \bibinfo{volume}{192}\/}, \bibinfo{pages}{116335}.
  \DOIprefix\doi{10.1016/j.eswa.2021.116335}.
\bibitem[{Arnab et~al.(2021)Arnab, Dehghani, Heigold, Sun, Lucic \&
  Schmid}]{vivit}
\bibinfo{author}{Arnab, A.}, \bibinfo{author}{Dehghani, M.},
  \bibinfo{author}{Heigold, G.}, \bibinfo{author}{Sun, C.},
  \bibinfo{author}{Lucic, M.}, \& \bibinfo{author}{Schmid, C.}
  (\bibinfo{year}{2021}).
\newblock \bibinfo{title}{{ViViT}: A video vision transformer}.
\newblock In {\it \bibinfo{booktitle}{2021 IEEE/CVF International Conference on
  Computer Vision, ICCV 2021, Montreal, QC, Canada, October 10-17, 2021}\/}
  (pp. \bibinfo{pages}{6816--6826}).
\newblock \bibinfo{publisher}{IEEE}.
\newblock \DOIprefix\doi{10.1109/ICCV48922.2021.00676}.
\bibitem[{Bain et~al.(2020)Bain, Nagrani, Brown \& Zisserman}]{condensed}
\bibinfo{author}{Bain, M.}, \bibinfo{author}{Nagrani, A.},
  \bibinfo{author}{Brown, A.}, \& \bibinfo{author}{Zisserman, A.}
  (\bibinfo{year}{2020}).
\newblock \bibinfo{title}{Condensed movies: Story based retrieval with
  contextual embeddings}.
\newblock In \bibinfo{editor}{H.~Ishikawa}, \bibinfo{editor}{C.~Liu},
  \bibinfo{editor}{T.~Pajdla}, \& \bibinfo{editor}{J.~Shi} (Eds.), {\it
  \bibinfo{booktitle}{Computer Vision - {ACCV} 2020 - 15th Asian Conference on
  Computer Vision, Kyoto, Japan, November 30 - December 4, 2020, Revised
  Selected Papers, Part {V}}\/} (pp. \bibinfo{pages}{460--479}).
\newblock \bibinfo{publisher}{Springer} volume \bibinfo{volume}{12626} of {\it
  \bibinfo{series}{Lecture Notes in Computer Science}\/}.
\newblock \DOIprefix\doi{10.1007/978-3-030-69541-5\_28}.
\bibitem[{Bernstein et~al.(2021)Bernstein, Sludds, Hamerly, Sze, Emer \&
  Englund}]{larger2}
\bibinfo{author}{Bernstein, L.}, \bibinfo{author}{Sludds, A.},
  \bibinfo{author}{Hamerly, R.}, \bibinfo{author}{Sze, V.},
  \bibinfo{author}{Emer, J.}, \& \bibinfo{author}{Englund, D.}
  (\bibinfo{year}{2021}).
\newblock \bibinfo{title}{Freely scalable and reconfigurable optical hardware
  for deep learning}.
\newblock {\it \bibinfo{journal}{Scientific reports}\/},  {\it
  \bibinfo{volume}{11}\/}, \bibinfo{pages}{3144}.
  \DOIprefix\doi{10.1038/s41598-021-82543-3}.
\bibitem[{Bose et~al.(2023)Bose, Hebbar, Somandepalli, Zhang, Cui,
  Cole{-}McLaughlin, Wang \& Narayanan}]{movieclip}
\bibinfo{author}{Bose, D.}, \bibinfo{author}{Hebbar, R.},
  \bibinfo{author}{Somandepalli, K.}, \bibinfo{author}{Zhang, H.},
  \bibinfo{author}{Cui, Y.}, \bibinfo{author}{Cole{-}McLaughlin, K.},
  \bibinfo{author}{Wang, H.}, \& \bibinfo{author}{Narayanan, S.}
  (\bibinfo{year}{2023}).
\newblock \bibinfo{title}{{MovieCLIP}: Visual scene recognition in movies}.
\newblock In {\it \bibinfo{booktitle}{{IEEE/CVF} Winter Conference on
  Applications of Computer Vision, {WACV} 2023, Waikoloa, HI, USA, January 2-7,
  2023}\/} (pp. \bibinfo{pages}{2082--2091}).
\newblock \bibinfo{publisher}{{IEEE}}.
\newblock \DOIprefix\doi{10.1109/WACV56688.2023.00212}.
\bibitem[{Bucila et~al.(2006)Bucila, Caruana \&
  Niculescu{-}Mizil}]{distillation_2006}
\bibinfo{author}{Bucila, C.}, \bibinfo{author}{Caruana, R.}, \&
  \bibinfo{author}{Niculescu{-}Mizil, A.} (\bibinfo{year}{2006}).
\newblock \bibinfo{title}{Model compression}.
\newblock In \bibinfo{editor}{T.~Eliassi{-}Rad}, \bibinfo{editor}{L.~H. Ungar},
  \bibinfo{editor}{M.~Craven}, \& \bibinfo{editor}{D.~Gunopulos} (Eds.), {\it
  \bibinfo{booktitle}{Proceedings of the Twelfth {ACM} {SIGKDD} International
  Conference on Knowledge Discovery and Data Mining, Philadelphia, PA, USA,
  August 20-23, 2006}\/} (pp. \bibinfo{pages}{535--541}).
\newblock \bibinfo{publisher}{{ACM}}.
\newblock \DOIprefix\doi{10.1145/1150402.1150464}.
\bibitem[{Carreira \& Zisserman(2017)}]{i3d}
\bibinfo{author}{Carreira, J.}, \& \bibinfo{author}{Zisserman, A.}
  (\bibinfo{year}{2017}).
\newblock \bibinfo{title}{Quo vadis, action recognition? a new model and the
  kinetics dataset}.
\newblock In {\it \bibinfo{booktitle}{2017 IEEE Conference on Computer Vision
  and Pattern Recognition, CVPR 2017, Honolulu, HI, USA, July 21-26, 2017}\/}
  (pp. \bibinfo{pages}{4724--4733}).
\newblock \bibinfo{publisher}{IEEE Computer Society}.
\newblock \DOIprefix\doi{10.1109/CVPR.2017.502}.
\bibitem[{{Cascante-Bonilla} et~al.(2019){Cascante-Bonilla}, Sitaraman, Luo \&
  Ordonez}]{moviescope}
\bibinfo{author}{{Cascante-Bonilla}, P.}, \bibinfo{author}{Sitaraman, K.},
  \bibinfo{author}{Luo, M.}, \& \bibinfo{author}{Ordonez, V.}
  (\bibinfo{year}{2019}).
\newblock \bibinfo{title}{Moviescope: Large-scale analysis of movies using
  multiple modalities}.
\newblock {\it \bibinfo{journal}{arXiv preprint arXiv:1908.03180}\/}, .
  \DOIprefix\doi{10.48550/arXiv.1908.03180}.
  \href{http://arxiv.org/abs/1908.03180}{\tt arXiv:1908.03180}.
\bibitem[{Choi et~al.(2016)Choi, Fazekas \& Sandler}]{musicnet}
\bibinfo{author}{Choi, K.}, \bibinfo{author}{Fazekas, G.}, \&
  \bibinfo{author}{Sandler, M.~B.} (\bibinfo{year}{2016}).
\newblock \bibinfo{title}{Automatic tagging using deep convolutional neural
  networks}.
\newblock In \bibinfo{editor}{M.~I. Mandel}, \bibinfo{editor}{J.~Devaney},
  \bibinfo{editor}{D.~Turnbull}, \& \bibinfo{editor}{G.~Tzanetakis} (Eds.),
  {\it \bibinfo{booktitle}{Proceedings of the 17th International Society for
  Music Information Retrieval Conference, {ISMIR} 2016, New York City, United
  States, August 7-11, 2016}\/} (pp. \bibinfo{pages}{805--811}).
\newblock \URLprefix
  \url{https://wp.nyu.edu/ismir2016/wp-content/uploads/sites/2294/2016/07/009\_Paper.pdf}.
\bibitem[{Defernez \& Kemsley(1999)}]{overfitting}
\bibinfo{author}{Defernez, M.}, \& \bibinfo{author}{Kemsley, E.~K.}
  (\bibinfo{year}{1999}).
\newblock \bibinfo{title}{Avoiding overfitting in the analysis of
  high-dimensional data with artificial neural networks ({ANNs})}.
\newblock {\it \bibinfo{journal}{Analyst}\/},  {\it \bibinfo{volume}{124}\/},
  \bibinfo{pages}{1675--1681}. \DOIprefix\doi{10.1039/A905556H}.
\bibitem[{Deldjoo et~al.(2018)Deldjoo, Constantin, Ionescu, Schedl \&
  Cremonesi}]{mmtf}
\bibinfo{author}{Deldjoo, Y.}, \bibinfo{author}{Constantin, M.~G.},
  \bibinfo{author}{Ionescu, B.}, \bibinfo{author}{Schedl, M.}, \&
  \bibinfo{author}{Cremonesi, P.} (\bibinfo{year}{2018}).
\newblock \bibinfo{title}{{MMTF-14K:} a multifaceted movie trailer feature
  dataset for recommendation and retrieval}.
\newblock In \bibinfo{editor}{P.~C{\'{e}}sar}, \bibinfo{editor}{M.~Zink}, \&
  \bibinfo{editor}{N.~Murray} (Eds.), {\it \bibinfo{booktitle}{Proceedings of
  the 9th {ACM} Multimedia Systems Conference, MMSys 2018, Amsterdam, The
  Netherlands, June 12-15, 2018}\/} (pp. \bibinfo{pages}{450--455}).
\newblock \bibinfo{publisher}{{ACM}}.
\newblock \DOIprefix\doi{10.1145/3204949.3208141}.
\bibitem[{Deng et~al.(2009)Deng, Dong, Socher, Li, Li \& Fei{-}Fei}]{imagenet}
\bibinfo{author}{Deng, J.}, \bibinfo{author}{Dong, W.},
  \bibinfo{author}{Socher, R.}, \bibinfo{author}{Li, L.}, \bibinfo{author}{Li,
  K.}, \& \bibinfo{author}{Fei{-}Fei, L.} (\bibinfo{year}{2009}).
\newblock \bibinfo{title}{{ImageNet}: {A} large-scale hierarchical image
  database}.
\newblock In {\it \bibinfo{booktitle}{2009 {IEEE} Computer Society Conference
  on Computer Vision and Pattern Recognition {(CVPR} 2009), 20-25 June 2009,
  Miami, Florida, {USA}}\/} (pp. \bibinfo{pages}{248--255}).
\newblock \bibinfo{publisher}{{IEEE} Computer Society}.
\newblock \DOIprefix\doi{10.1109/CVPR.2009.5206848}.
\bibitem[{Devlin et~al.(2019)Devlin, Chang, Lee \& Toutanova}]{bert}
\bibinfo{author}{Devlin, J.}, \bibinfo{author}{Chang, M.},
  \bibinfo{author}{Lee, K.}, \& \bibinfo{author}{Toutanova, K.}
  (\bibinfo{year}{2019}).
\newblock \bibinfo{title}{{BERT:} pre-training of deep bidirectional
  transformers for language understanding}.
\newblock In \bibinfo{editor}{J.~Burstein}, \bibinfo{editor}{C.~Doran}, \&
  \bibinfo{editor}{T.~Solorio} (Eds.), {\it \bibinfo{booktitle}{Proceedings of
  the 2019 Conference of the North American Chapter of the Association for
  Computational Linguistics: Human Language Technologies, {NAACL-HLT} 2019,
  Minneapolis, MN, USA, June 2-7, 2019, Volume 1 (Long and Short Papers)}\/}
  (pp. \bibinfo{pages}{4171--4186}).
\newblock \bibinfo{publisher}{Association for Computational Linguistics}.
\newblock \DOIprefix\doi{10.18653/v1/n19-1423}.
\bibitem[{Dosovitskiy et~al.(2021)Dosovitskiy, Beyer, Kolesnikov, Weissenborn,
  Zhai, Unterthiner, Dehghani, Minderer, Heigold, Gelly, Uszkoreit \&
  Houlsby}]{vit}
\bibinfo{author}{Dosovitskiy, A.}, \bibinfo{author}{Beyer, L.},
  \bibinfo{author}{Kolesnikov, A.}, \bibinfo{author}{Weissenborn, D.},
  \bibinfo{author}{Zhai, X.}, \bibinfo{author}{Unterthiner, T.},
  \bibinfo{author}{Dehghani, M.}, \bibinfo{author}{Minderer, M.},
  \bibinfo{author}{Heigold, G.}, \bibinfo{author}{Gelly, S.},
  \bibinfo{author}{Uszkoreit, J.}, \& \bibinfo{author}{Houlsby, N.}
  (\bibinfo{year}{2021}).
\newblock \bibinfo{title}{An image is worth 16x16 words: Transformers for image
  recognition at scale}.
\newblock In {\it \bibinfo{booktitle}{9th International Conference on Learning
  Representations, ICLR 2021, Virtual Event, Austria, May 3-7, 2021}\/}.
\newblock \bibinfo{publisher}{OpenReview.net}.
\newblock \URLprefix \url{https://openreview.net/forum?id=YicbFdNTTy}.
\bibitem[{Gemmeke et~al.(2017)Gemmeke, Ellis, Freedman, Jansen, Lawrence,
  Moore, Plakal \& Ritter}]{audioset}
\bibinfo{author}{Gemmeke, J.~F.}, \bibinfo{author}{Ellis, D. P.~W.},
  \bibinfo{author}{Freedman, D.}, \bibinfo{author}{Jansen, A.},
  \bibinfo{author}{Lawrence, W.}, \bibinfo{author}{Moore, R.~C.},
  \bibinfo{author}{Plakal, M.}, \& \bibinfo{author}{Ritter, M.}
  (\bibinfo{year}{2017}).
\newblock \bibinfo{title}{Audio set: {An} ontology and human-labeled dataset
  for audio events}.
\newblock In {\it \bibinfo{booktitle}{2017 IEEE International Conference on
  Acoustics, Speech and Signal Processing, ICASSP 2017, New Orleans, LA, USA,
  March 5-9, 2017}\/} (pp. \bibinfo{pages}{776--780}).
\newblock \bibinfo{publisher}{IEEE}.
\newblock \DOIprefix\doi{10.1109/ICASSP.2017.7952261}.
\bibitem[{Gong et~al.(2021)Gong, Chung \& Glass}]{ast}
\bibinfo{author}{Gong, Y.}, \bibinfo{author}{Chung, Y.-A.}, \&
  \bibinfo{author}{Glass, J.~R.} (\bibinfo{year}{2021}).
\newblock \bibinfo{title}{{AST}: Audio spectrogram transformer}.
\newblock In \bibinfo{editor}{H.~Hermansky}, \bibinfo{editor}{H.~Cernock{\'y}},
  \bibinfo{editor}{L.~Burget}, \bibinfo{editor}{L.~Lamel},
  \bibinfo{editor}{O.~Scharenborg}, \& \bibinfo{editor}{P.~Motl{\'i}cek}
  (Eds.), {\it \bibinfo{booktitle}{Interspeech 2021, 22nd Annual Conference of
  the International Speech Communication Association, Brno, Czechia, 30 August
  - 3 September 2021}\/} (pp. \bibinfo{pages}{571--575}).
\newblock \bibinfo{publisher}{ISCA}.
\newblock \DOIprefix\doi{10.21437/Interspeech.2021-698}.
\bibitem[{Guhr(2023)}]{spell_correction}
\bibinfo{author}{Guhr, O.} (\bibinfo{year}{2023}).
\newblock \bibinfo{title}{Spelling-correction-english-base}.
\newblock
  \bibinfo{howpublished}{\url{https://huggingface.co/oliverguhr/spelling-correction-english-base}}.
\newblock \bibinfo{note}{Accessed: 2023-07-03}.
\bibitem[{Harper \& Konstan(2016)}]{movielens}
\bibinfo{author}{Harper, F.~M.}, \& \bibinfo{author}{Konstan, J.~A.}
  (\bibinfo{year}{2016}).
\newblock \bibinfo{title}{The {MovieLens} datasets: History and context}.
\newblock {\it \bibinfo{journal}{ACM Trans. Interact. Intell. Syst.}\/},  {\it
  \bibinfo{volume}{5}\/}, \bibinfo{pages}{19:1--19:19}.
  \DOIprefix\doi{10.1145/2827872}.
\bibitem[{Hinton et~al.(2015)Hinton, Vinyals \& Dean}]{distillation_2015}
\bibinfo{author}{Hinton, G.}, \bibinfo{author}{Vinyals, O.}, \&
  \bibinfo{author}{Dean, J.} (\bibinfo{year}{2015}).
\newblock \bibinfo{title}{Distilling the knowledge in a neural network}.
\newblock {\it \bibinfo{journal}{arXiv preprint arXiv:1503.02531}\/}, .
  \DOIprefix\doi{10.48550/arXiv.1503.02531}.
\bibitem[{Huang et~al.(2020)Huang, Xiong, Rao, Wang \& Lin}]{movienet}
\bibinfo{author}{Huang, Q.}, \bibinfo{author}{Xiong, Y.}, \bibinfo{author}{Rao,
  A.}, \bibinfo{author}{Wang, J.}, \& \bibinfo{author}{Lin, D.}
  (\bibinfo{year}{2020}).
\newblock \bibinfo{title}{{MovieNet}: A holistic dataset for movie
  understanding}.
\newblock In \bibinfo{editor}{A.~Vedaldi}, \bibinfo{editor}{H.~Bischof},
  \bibinfo{editor}{T.~Brox}, \& \bibinfo{editor}{J.-M. Frahm} (Eds.), {\it
  \bibinfo{booktitle}{Computer Vision - ECCV 2020 - 16th European Conference,
  Glasgow, UK, August 23-28, 2020, Proceedings, Part IV}\/} (pp.
  \bibinfo{pages}{709--727}).
\newblock \bibinfo{publisher}{Springer} volume \bibinfo{volume}{12349} of {\it
  \bibinfo{series}{Lecture Notes in Computer Science}\/}.
\newblock \DOIprefix\doi{10.1007/978-3-030-58548-8_41}.
\bibitem[{Khan et~al.(2024)Khan, Hussain, Ullah~Khan, Ahmad~Khan \&
  Baik}]{summarization_e}
\bibinfo{author}{Khan, H.}, \bibinfo{author}{Hussain, T.},
  \bibinfo{author}{Ullah~Khan, S.}, \bibinfo{author}{Ahmad~Khan, Z.}, \&
  \bibinfo{author}{Baik, S.~W.} (\bibinfo{year}{2024}).
\newblock \bibinfo{title}{Deep multi-scale pyramidal features network for
  supervised video summarization}.
\newblock {\it \bibinfo{journal}{Expert Systems with Applications}\/},  {\it
  \bibinfo{volume}{237}\/}, \bibinfo{pages}{121288}.
  \DOIprefix\doi{10.1016/j.eswa.2023.121288}.
\bibitem[{Kingma \& Ba(2015)}]{adam}
\bibinfo{author}{Kingma, D.~P.}, \& \bibinfo{author}{Ba, J.}
  (\bibinfo{year}{2015}).
\newblock \bibinfo{title}{Adam: {A} method for stochastic optimization}.
\newblock In \bibinfo{editor}{Y.~Bengio}, \& \bibinfo{editor}{Y.~LeCun} (Eds.),
  {\it \bibinfo{booktitle}{3rd International Conference on Learning
  Representations, {ICLR} 2015, San Diego, CA, USA, May 7-9, 2015, Conference
  Track Proceedings}\/}.
\newblock \URLprefix \url{http://arxiv.org/abs/1412.6980}.
\bibitem[{Kong et~al.(2020)Kong, Cao, Iqbal, Wang, Wang \& Plumbley}]{audiotag}
\bibinfo{author}{Kong, Q.}, \bibinfo{author}{Cao, Y.}, \bibinfo{author}{Iqbal,
  T.}, \bibinfo{author}{Wang, Y.}, \bibinfo{author}{Wang, W.}, \&
  \bibinfo{author}{Plumbley, M.~D.} (\bibinfo{year}{2020}).
\newblock \bibinfo{title}{{PANNs}: Large-scale pretrained audio neural networks
  for audio pattern recognition}.
\newblock {\it \bibinfo{journal}{{IEEE} {ACM} Trans. Audio Speech Lang.
  Process.}\/},  {\it \bibinfo{volume}{28}\/}, \bibinfo{pages}{2880--2894}.
  \DOIprefix\doi{10.1109/TASLP.2020.3030497}.
\bibitem[{Miyazawa et~al.(2022)Miyazawa, Kyuragi \&
  Nagai}]{trailer_poster_classification_2}
\bibinfo{author}{Miyazawa, K.}, \bibinfo{author}{Kyuragi, Y.}, \&
  \bibinfo{author}{Nagai, T.} (\bibinfo{year}{2022}).
\newblock \bibinfo{title}{Simple and effective multimodal learning based on
  pre-trained transformer models}.
\newblock {\it \bibinfo{journal}{IEEE Access}\/},  {\it
  \bibinfo{volume}{10}\/}, \bibinfo{pages}{29821--29833}.
  \DOIprefix\doi{10.1109/ACCESS.2022.3159346}.
\bibitem[{Niu et~al.(2020)Niu, Liu, Wang \& Song}]{transfer2}
\bibinfo{author}{Niu, S.}, \bibinfo{author}{Liu, Y.}, \bibinfo{author}{Wang,
  J.}, \& \bibinfo{author}{Song, H.} (\bibinfo{year}{2020}).
\newblock \bibinfo{title}{A decade survey of transfer learning (2010--2020)}.
\newblock {\it \bibinfo{journal}{IEEE Transactions on Artificial
  Intelligence}\/},  {\it \bibinfo{volume}{1}\/}, \bibinfo{pages}{151--166}.
  \DOIprefix\doi{10.1109/TAI.2021.3054609}.
\bibitem[{Ovalle et~al.(2017)Ovalle, Solorio, {Montes-y-G{\'o}mez} \&
  Gonz{\'a}lez}]{mm-imdb}
\bibinfo{author}{Ovalle, J. E.~A.}, \bibinfo{author}{Solorio, T.},
  \bibinfo{author}{{Montes-y-G{\'o}mez}, M.}, \& \bibinfo{author}{Gonz{\'a}lez,
  F.~A.} (\bibinfo{year}{2017}).
\newblock \bibinfo{title}{Gated multimodal units for information fusion}.
\newblock In {\it \bibinfo{booktitle}{5th International Conference on Learning
  Representations, ICLR 2017, Toulon, France, April 24-26, 2017, Workshop Track
  Proceedings}\/}.
\newblock \bibinfo{publisher}{OpenReview.net}.
\newblock \URLprefix \url{https://openreview.net/forum?id=B1akgy9xx}.
\bibitem[{PaddlePaddle(2023)}]{paddle}
\bibinfo{author}{PaddlePaddle} (\bibinfo{year}{2023}).
\newblock \bibinfo{title}{{PaddleOCR}}.
\newblock
  \bibinfo{howpublished}{\url{https://github.com/PaddlePaddle/PaddleOCR}}.
\newblock \bibinfo{note}{Accessed: 2023-07-03}.
\bibitem[{Papariello(2022)}]{language_detection}
\bibinfo{author}{Papariello, L.} (\bibinfo{year}{2022}).
\newblock \bibinfo{title}{xlm-roberta-base-language-detection}.
\newblock
  \bibinfo{howpublished}{\url{https://huggingface.co/papluca/xlm-roberta-base-language-detection}}.
\newblock \bibinfo{note}{Accessed: 2023-07-03}.
\bibitem[{Paszke et~al.(2019)Paszke, Gross, Massa, Lerer, Bradbury, Chanan,
  Killeen, Lin, Gimelshein, Antiga, Desmaison, K{\"o}pf, Yang, DeVito, Raison,
  Tejani, Chilamkurthy, Steiner, Fang, Bai \& Chintala}]{pytorch}
\bibinfo{author}{Paszke, A.}, \bibinfo{author}{Gross, S.},
  \bibinfo{author}{Massa, F.}, \bibinfo{author}{Lerer, A.},
  \bibinfo{author}{Bradbury, J.}, \bibinfo{author}{Chanan, G.},
  \bibinfo{author}{Killeen, T.}, \bibinfo{author}{Lin, Z.},
  \bibinfo{author}{Gimelshein, N.}, \bibinfo{author}{Antiga, L.},
  \bibinfo{author}{Desmaison, A.}, \bibinfo{author}{K{\"o}pf, A.},
  \bibinfo{author}{Yang, E.}, \bibinfo{author}{DeVito, Z.},
  \bibinfo{author}{Raison, M.}, \bibinfo{author}{Tejani, A.},
  \bibinfo{author}{Chilamkurthy, S.}, \bibinfo{author}{Steiner, B.},
  \bibinfo{author}{Fang, L.}, \bibinfo{author}{Bai, J.}, \&
  \bibinfo{author}{Chintala, S.} (\bibinfo{year}{2019}).
\newblock \bibinfo{title}{{PyTorch}: An imperative style, high-performance deep
  learning library}.
\newblock In \bibinfo{editor}{H.~M. Wallach}, \bibinfo{editor}{H.~Larochelle},
  \bibinfo{editor}{A.~Beygelzimer}, \bibinfo{editor}{F.~{d'Alch{\'e}-Buc}},
  \bibinfo{editor}{E.~B. Fox}, \& \bibinfo{editor}{R.~Garnett} (Eds.), {\it
  \bibinfo{booktitle}{Advances in Neural Information Processing Systems 32:
  Annual Conference on Neural Information Processing Systems 2019, NeurIPS
  2019, 8-14 December 2019, Vancouver, BC, Canada}\/} (pp.
  \bibinfo{pages}{8024--8035}).
\newblock \URLprefix
  \url{https://proceedings.neurips.cc/paper_files/paper/2019/file/bdbca288fee7f92f2bfa9f7012727740-Paper.pdf}.
\bibitem[{Radford et~al.(2021)Radford, Kim, Hallacy, Ramesh, Goh, Agarwal,
  Sastry, Askell, Mishkin, Clark, Krueger \& Sutskever}]{clip}
\bibinfo{author}{Radford, A.}, \bibinfo{author}{Kim, J.~W.},
  \bibinfo{author}{Hallacy, C.}, \bibinfo{author}{Ramesh, A.},
  \bibinfo{author}{Goh, G.}, \bibinfo{author}{Agarwal, S.},
  \bibinfo{author}{Sastry, G.}, \bibinfo{author}{Askell, A.},
  \bibinfo{author}{Mishkin, P.}, \bibinfo{author}{Clark, J.},
  \bibinfo{author}{Krueger, G.}, \& \bibinfo{author}{Sutskever, I.}
  (\bibinfo{year}{2021}).
\newblock \bibinfo{title}{Learning transferable visual models from natural
  language supervision}.
\newblock In \bibinfo{editor}{M.~Meila}, \& \bibinfo{editor}{T.~Zhang} (Eds.),
  {\it \bibinfo{booktitle}{Proceedings of the 38th International Conference on
  Machine Learning, {ICML} 2021, 18-24 July 2021, Virtual Event}\/} (pp.
  \bibinfo{pages}{8748--8763}).
\newblock \bibinfo{publisher}{{PMLR}} volume \bibinfo{volume}{139} of {\it
  \bibinfo{series}{Proceedings of Machine Learning Research}\/}.
\newblock \URLprefix \url{http://proceedings.mlr.press/v139/radford21a.html}.
\bibitem[{Radford et~al.(2023)Radford, Kim, Xu, Brockman, McLeavey \&
  Sutskever}]{whisper}
\bibinfo{author}{Radford, A.}, \bibinfo{author}{Kim, J.~W.},
  \bibinfo{author}{Xu, T.}, \bibinfo{author}{Brockman, G.},
  \bibinfo{author}{McLeavey, C.}, \& \bibinfo{author}{Sutskever, I.}
  (\bibinfo{year}{2023}).
\newblock \bibinfo{title}{Robust speech recognition via large-scale weak
  supervision}.
\newblock In \bibinfo{editor}{A.~Krause}, \bibinfo{editor}{E.~Brunskill},
  \bibinfo{editor}{K.~Cho}, \bibinfo{editor}{B.~Engelhardt},
  \bibinfo{editor}{S.~Sabato}, \& \bibinfo{editor}{J.~Scarlett} (Eds.), {\it
  \bibinfo{booktitle}{International Conference on Machine Learning, ICML 2023,
  23-29 July 2023, Honolulu, Hawaii, USA}\/} (pp.
  \bibinfo{pages}{28492--28518}).
\newblock \bibinfo{publisher}{PMLR} volume \bibinfo{volume}{202} of {\it
  \bibinfo{series}{Proceedings of Machine Learning Research}\/}.
\newblock \URLprefix \url{https://proceedings.mlr.press/v202/radford23a.html}.
\bibitem[{Ray \& Kolekar(2024)}]{activity_e}
\bibinfo{author}{Ray, A.}, \& \bibinfo{author}{Kolekar, M.~H.}
  (\bibinfo{year}{2024}).
\newblock \bibinfo{title}{Transfer learning and its extensive appositeness in
  human activity recognition: A survey}.
\newblock {\it \bibinfo{journal}{Expert Systems with Applications}\/},  {\it
  \bibinfo{volume}{240}\/}, \bibinfo{pages}{122538}.
  \DOIprefix\doi{10.1016/j.eswa.2023.122538}.
\bibitem[{Rodr{\'{\i}}guez~Bribiesca et~al.(2021)Rodr{\'{\i}}guez~Bribiesca,
  L{\'o}pez~Monroy \& {Montes-y-G{\'o}mez}}]{trailer_metadata_classification}
\bibinfo{author}{Rodr{\'{\i}}guez~Bribiesca, I.},
  \bibinfo{author}{L{\'o}pez~Monroy, A.~P.}, \&
  \bibinfo{author}{{Montes-y-G{\'o}mez}, M.} (\bibinfo{year}{2021}).
\newblock \bibinfo{title}{Multimodal weighted fusion of transformers for movie
  genre classification}.
\newblock In \bibinfo{editor}{A.~Zadeh}, \bibinfo{editor}{L.-P. Morency},
  \bibinfo{editor}{P.~P. Liang}, \bibinfo{editor}{C.~Ross},
  \bibinfo{editor}{R.~Salakhutdinov}, \bibinfo{editor}{S.~Poria},
  \bibinfo{editor}{E.~Cambria}, \& \bibinfo{editor}{K.~Shi} (Eds.), {\it
  \bibinfo{booktitle}{Proceedings of the Third Workshop on Multimodal
  Artificial Intelligence}\/} (pp. \bibinfo{pages}{1--5}).
\newblock \bibinfo{address}{Mexico City, Mexico}:
  \bibinfo{publisher}{Association for Computational Linguistics}.
\newblock \DOIprefix\doi{10.18653/v1/2021.maiworkshop-1.1}.
\bibitem[{Sanh et~al.(2019)Sanh, Debut, Chaumond \& Wolf}]{distilbert}
\bibinfo{author}{Sanh, V.}, \bibinfo{author}{Debut, L.},
  \bibinfo{author}{Chaumond, J.}, \& \bibinfo{author}{Wolf, T.}
  (\bibinfo{year}{2019}).
\newblock \bibinfo{title}{{DistilBERT}, a distilled version of {BERT}: smaller,
  faster, cheaper and lighter}.
\newblock {\it \bibinfo{journal}{arXiv preprint arXiv:1910.01108}\/}, .
  \DOIprefix\doi{10.48550/arXiv.1910.01108}.
\bibitem[{Simoes et~al.(2016)Simoes, Wehrmann, Barros \& Ruiz}]{lmtd}
\bibinfo{author}{Simoes, G.~S.}, \bibinfo{author}{Wehrmann, J.},
  \bibinfo{author}{Barros, R.~C.}, \& \bibinfo{author}{Ruiz, D.~D.}
  (\bibinfo{year}{2016}).
\newblock \bibinfo{title}{Movie genre classification with convolutional neural
  networks}.
\newblock In {\it \bibinfo{booktitle}{2016 International Joint Conference on
  Neural Networks, {IJCNN} 2016, Vancouver, BC, Canada, July 24-29, 2016}\/}
  (pp. \bibinfo{pages}{259--266}).
\newblock \bibinfo{publisher}{{IEEE}}.
\newblock \DOIprefix\doi{10.1109/IJCNN.2016.7727207}.
\bibitem[{Srivastava et~al.(2014)Srivastava, Hinton, Krizhevsky, Sutskever \&
  Salakhutdinov}]{dropout}
\bibinfo{author}{Srivastava, N.}, \bibinfo{author}{Hinton, G.},
  \bibinfo{author}{Krizhevsky, A.}, \bibinfo{author}{Sutskever, I.}, \&
  \bibinfo{author}{Salakhutdinov, R.} (\bibinfo{year}{2014}).
\newblock \bibinfo{title}{Dropout: {A} simple way to prevent neural networks
  from overfitting}.
\newblock {\it \bibinfo{journal}{The journal of machine learning research}\/},
  {\it \bibinfo{volume}{15}\/}, \bibinfo{pages}{1929--1958}.
  \DOIprefix\doi{10.5555/2627435.2670313}.
\bibitem[{Tan et~al.(2018)Tan, Sun, Kong, Zhang, Yang \& Liu}]{transfer1}
\bibinfo{author}{Tan, C.}, \bibinfo{author}{Sun, F.}, \bibinfo{author}{Kong,
  T.}, \bibinfo{author}{Zhang, W.}, \bibinfo{author}{Yang, C.}, \&
  \bibinfo{author}{Liu, C.} (\bibinfo{year}{2018}).
\newblock \bibinfo{title}{A survey on deep transfer learning}.
\newblock In \bibinfo{editor}{V.~K{\r u}rkov{\'a}},
  \bibinfo{editor}{Y.~Manolopoulos}, \bibinfo{editor}{B.~Hammer},
  \bibinfo{editor}{L.~Iliadis}, \& \bibinfo{editor}{I.~Maglogiannis} (Eds.),
  {\it \bibinfo{booktitle}{Artificial Neural Networks and Machine Learning --
  ICANN 2018}\/} (pp. \bibinfo{pages}{270--279}).
\newblock \bibinfo{address}{Cham}: \bibinfo{publisher}{Springer International
  Publishing} volume \bibinfo{volume}{11141}.
\newblock \DOIprefix\doi{10.1007/978-3-030-01424-7_27}.
\bibitem[{Tan \& Le(2019)}]{larger1}
\bibinfo{author}{Tan, M.}, \& \bibinfo{author}{Le, Q.} (\bibinfo{year}{2019}).
\newblock \bibinfo{title}{{E}fficient{N}et: Rethinking model scaling for
  convolutional neural networks}.
\newblock In \bibinfo{editor}{K.~Chaudhuri}, \&
  \bibinfo{editor}{R.~Salakhutdinov} (Eds.), {\it
  \bibinfo{booktitle}{Proceedings of the 36th International Conference on
  Machine Learning}\/} (pp. \bibinfo{pages}{6105--6114}).
\newblock \bibinfo{publisher}{PMLR} volume~\bibinfo{volume}{97} of {\it
  \bibinfo{series}{Proceedings of Machine Learning Research}\/}.
\newblock \URLprefix \url{https://proceedings.mlr.press/v97/tan19a.html}.
\bibitem[{Tomar(2006)}]{ffmpeg}
\bibinfo{author}{Tomar, S.} (\bibinfo{year}{2006}).
\newblock \bibinfo{title}{Converting video formats with ffmpeg}.
\newblock {\it \bibinfo{journal}{Linux journal}\/},  {\it
  \bibinfo{volume}{2006}\/}, \bibinfo{pages}{10}. \URLprefix
  \url{https://dl.acm.org/doi/abs/10.5555/1134782.1134792}.
\bibitem[{Vaswani et~al.(2017)Vaswani, Shazeer, Parmar, Uszkoreit, Jones,
  Gomez, Kaiser \& Polosukhin}]{transformer}
\bibinfo{author}{Vaswani, A.}, \bibinfo{author}{Shazeer, N.},
  \bibinfo{author}{Parmar, N.}, \bibinfo{author}{Uszkoreit, J.},
  \bibinfo{author}{Jones, L.}, \bibinfo{author}{Gomez, A.~N.},
  \bibinfo{author}{Kaiser, L.}, \& \bibinfo{author}{Polosukhin, I.}
  (\bibinfo{year}{2017}).
\newblock \bibinfo{title}{Attention is all you need}.
\newblock In \bibinfo{editor}{I.~Guyon}, \bibinfo{editor}{U.~V. Luxburg},
  \bibinfo{editor}{S.~Bengio}, \bibinfo{editor}{H.~Wallach},
  \bibinfo{editor}{R.~Fergus}, \bibinfo{editor}{S.~Vishwanathan}, \&
  \bibinfo{editor}{R.~Garnett} (Eds.), {\it \bibinfo{booktitle}{Advances in
  Neural Information Processing Systems 30: Annual Conference on Neural
  Information Processing Systems}\/} (pp. \bibinfo{pages}{5998--6008}).
\newblock \URLprefix
  \url{https://proceedings.neurips.cc/paper/2017/hash/3f5ee243547dee91fbd053c1c4a845aa-Abstract.html}.
\bibitem[{Wang et~al.(2020)Wang, Zhao, Lioma, Li, Zhang \&
  Simonsen}]{learned_position}
\bibinfo{author}{Wang, B.}, \bibinfo{author}{Zhao, D.}, \bibinfo{author}{Lioma,
  C.}, \bibinfo{author}{Li, Q.}, \bibinfo{author}{Zhang, P.}, \&
  \bibinfo{author}{Simonsen, J.~G.} (\bibinfo{year}{2020}).
\newblock \bibinfo{title}{Encoding word order in complex embeddings}.
\newblock In {\it \bibinfo{booktitle}{8th International Conference on Learning
  Representations, {ICLR} 2020, Addis Ababa, Ethiopia, April 26-30, 2020}\/}.
\newblock \bibinfo{publisher}{OpenReview.net}.
\newblock \URLprefix \url{https://openreview.net/forum?id=Hke-WTVtwr}.
\bibitem[{Wang et~al.(2016)Wang, Xiong, Wang, Qiao, Lin, Tang \& Gool}]{tsn}
\bibinfo{author}{Wang, L.}, \bibinfo{author}{Xiong, Y.}, \bibinfo{author}{Wang,
  Z.}, \bibinfo{author}{Qiao, Y.}, \bibinfo{author}{Lin, D.},
  \bibinfo{author}{Tang, X.}, \& \bibinfo{author}{Gool, L.~V.}
  (\bibinfo{year}{2016}).
\newblock \bibinfo{title}{Temporal segment networks: Towards good practices for
  deep action recognition}.
\newblock In \bibinfo{editor}{B.~Leibe}, \bibinfo{editor}{J.~Matas},
  \bibinfo{editor}{N.~Sebe}, \& \bibinfo{editor}{M.~Welling} (Eds.), {\it
  \bibinfo{booktitle}{Computer Vision - ECCV 2016 - 14th European Conference,
  Amsterdam, The Netherlands, October 11-14, 2016, Proceedings, Part VIII}\/}
  (pp. \bibinfo{pages}{20--36}).
\newblock \bibinfo{publisher}{Springer} volume \bibinfo{volume}{9912} of {\it
  \bibinfo{series}{Lecture Notes in Computer Science}\/}.
\newblock \DOIprefix\doi{10.1007/978-3-319-46484-8_2}.
\bibitem[{Wehrmann \& Barros(2017)}]{wehrmann}
\bibinfo{author}{Wehrmann, J.}, \& \bibinfo{author}{Barros, R.~C.}
  (\bibinfo{year}{2017}).
\newblock \bibinfo{title}{Movie genre classification: {A} multi-label approach
  based on convolutions through time}.
\newblock {\it \bibinfo{journal}{Appl. Soft Comput.}\/},  {\it
  \bibinfo{volume}{61}\/}, \bibinfo{pages}{973--982}.
  \DOIprefix\doi{10.1016/j.asoc.2017.08.029}.
\bibitem[{Yadav \& Vishwakarma(2020)}]{yadav}
\bibinfo{author}{Yadav, A.}, \& \bibinfo{author}{Vishwakarma, D.~K.}
  (\bibinfo{year}{2020}).
\newblock \bibinfo{title}{A unified framework of deep networks for genre
  classification using movie trailer}.
\newblock {\it \bibinfo{journal}{Appl. Soft Comput.}\/},  {\it
  \bibinfo{volume}{96}\/}, \bibinfo{pages}{106624}.
  \DOIprefix\doi{10.1016/j.asoc.2020.106624}.
\bibitem[{Zhou et~al.(2018{\natexlab{a}})Zhou, Andonian, Oliva \&
  Torralba}]{trn}
\bibinfo{author}{Zhou, B.}, \bibinfo{author}{Andonian, A.},
  \bibinfo{author}{Oliva, A.}, \& \bibinfo{author}{Torralba, A.}
  (\bibinfo{year}{2018}{\natexlab{a}}).
\newblock \bibinfo{title}{Temporal relational reasoning in videos}.
\newblock In \bibinfo{editor}{V.~Ferrari}, \bibinfo{editor}{M.~Hebert},
  \bibinfo{editor}{C.~Sminchisescu}, \& \bibinfo{editor}{Y.~Weiss} (Eds.), {\it
  \bibinfo{booktitle}{Computer Vision - ECCV 2018 - 15th European Conference,
  Munich, Germany, September 8-14, 2018, Proceedings, Part I}\/} (pp.
  \bibinfo{pages}{831--846}).
\newblock \bibinfo{publisher}{Springer} volume \bibinfo{volume}{11205} of {\it
  \bibinfo{series}{Lecture Notes in Computer Science}\/}.
\newblock \DOIprefix\doi{10.1007/978-3-030-01246-5_49}.
\bibitem[{Zhou et~al.(2018{\natexlab{b}})Zhou, Lapedriza, Khosla, Oliva \&
  Torralba}]{places}
\bibinfo{author}{Zhou, B.}, \bibinfo{author}{Lapedriza, {\`{A}}.},
  \bibinfo{author}{Khosla, A.}, \bibinfo{author}{Oliva, A.}, \&
  \bibinfo{author}{Torralba, A.} (\bibinfo{year}{2018}{\natexlab{b}}).
\newblock \bibinfo{title}{Places: {A} 10 million image database for scene
  recognition}.
\newblock {\it \bibinfo{journal}{{IEEE} Trans. Pattern Anal. Mach. Intell.}\/},
   {\it \bibinfo{volume}{40}\/}, \bibinfo{pages}{1452--1464}.
  \DOIprefix\doi{10.1109/TPAMI.2017.2723009}.
\bibitem[{Zhou et~al.(2010)Zhou, Hermans, Karandikar \& Rehg}]{zhou}
\bibinfo{author}{Zhou, H.}, \bibinfo{author}{Hermans, T.},
  \bibinfo{author}{Karandikar, A.~V.}, \& \bibinfo{author}{Rehg, J.~M.}
  (\bibinfo{year}{2010}).
\newblock \bibinfo{title}{Movie genre classification via scene categorization}.
\newblock In \bibinfo{editor}{A.~D. Bimbo}, \bibinfo{editor}{S.~Chang}, \&
  \bibinfo{editor}{A.~W.~M. Smeulders} (Eds.), {\it
  \bibinfo{booktitle}{Proceedings of the 18th International Conference on
  Multimedia 2010, Firenze, Italy, October 25-29, 2010}\/} (pp.
  \bibinfo{pages}{747--750}).
\newblock \bibinfo{publisher}{{ACM}}.
\newblock \DOIprefix\doi{10.1145/1873951.1874068}.

\end{thebibliography}

\end{document}